%% file: main.tex
\documentclass{article}
\usepackage{spconf,amsmath,graphicx}
\usepackage{threeparttable}
\usepackage{amsfonts}
\usepackage{booktabs}
\usepackage{url}
\usepackage{multirow}
\usepackage{amssymb}
\usepackage{bbding}
\usepackage{color}

\newcommand{\eg}{\emph{e.g.,}~}
\newcommand{\etal}{\emph{et al.}~}

\title{Hierarchical Similarity Learning for Language-based\\ Product Image Retrieval}

\name{Zhe Ma\textsuperscript{\rm 1},
Fenghao Liu\textsuperscript{\rm 1}, 
Jianfeng Dong\textsuperscript{\rm 2,\rm 3}\sthanks{Corresponding author: Jianfeng Dong},
Xiaoye Qu\textsuperscript{\rm 4},
Yuan He\textsuperscript{\rm 5},
Shouling Ji\textsuperscript{\rm 1,\rm 3}}
\address{\textsuperscript{\rm 1}Zhejiang University,
\textsuperscript{\rm 2}Zhejiang Gongshang University,\\
\textsuperscript{\rm 3}Alibaba-Zhejiang University Joint Research Institute of Frontier Technologies,\\
\textsuperscript{\rm 4}Huazhong University of Science and Technology,
\textsuperscript{\rm 5}Alibaba Group}

\begin{document}
\ninept

\maketitle

\begin{abstract}
This paper aims for the language-based product image retrieval task. The majority of previous works have made significant progress by designing network structure, similarity measurement, and loss function. However, they typically perform vision-text matching at certain granularity regardless of the intrinsic multiple granularities of images. In this paper, we focus on the cross-modal similarity measurement, and propose a novel \textbf{H}ierarchical \textbf{S}imilarity \textbf{L}earning (HSL) network. HSL first learns multi-level representations of input data by stacked encoders, and object-granularity similarity and image-granularity similarity are computed at each level. All the similarities are combined as the final hierarchical cross-modal similarity.
Experiments on a large-scale product retrieval dataset demonstrate the effectiveness of our proposed method. Code and data are available at https://github.com/liufh1/hsl.
\end{abstract}


\input{tex/intro}

\input{tex/method}

\input{tex/exp}

\input{tex/conclusion}

\section{Acknowledgments}
This work was partly supported by the National Key Research and Development Program of China under No. 2018YFB0804102 and No. 2020YFB2103802, NSFC under No. 61902347, No. 61772466, U1936215, and U1836202, the Zhejiang Provincial Natural Science Foundation under No. LQ19F020002 and No. LR19F020003, the Zhejiang Provincial Public Welfare Technology Research Project under No. LGF21F020010 and LGG20F020005, Alibaba-Zhejiang University Joint Research Institute of Frontier Technologies, and the Fundamental Research Funds for the Central Universities (Zhejiang University NGICS Platform).

\bibliographystyle{IEEE}
\bibliography{ref}

\end{document}

%% file: tex/intro.tex
\section{Introduction}
\begin{figure}[!t]
\begin{minipage}[b]{1.0\linewidth}
  \centering
  \centerline{\includegraphics[width=0.9\linewidth]{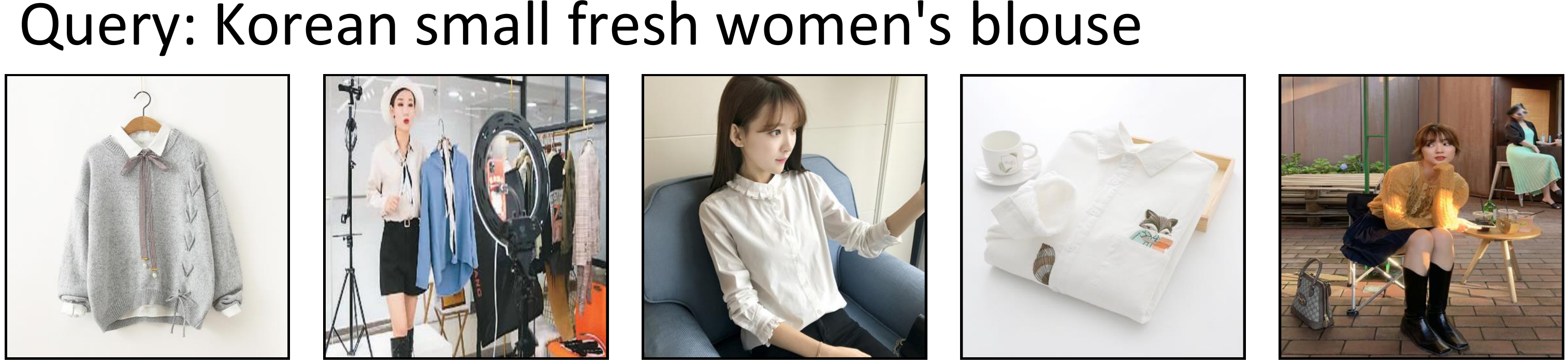}}\label{fig:concept_a} 
  \centerline{(a)}
\end{minipage}
 \hfill
\begin{minipage}[b]{1.0\linewidth}
  \centering
  \centerline{\includegraphics[width=0.99\linewidth]{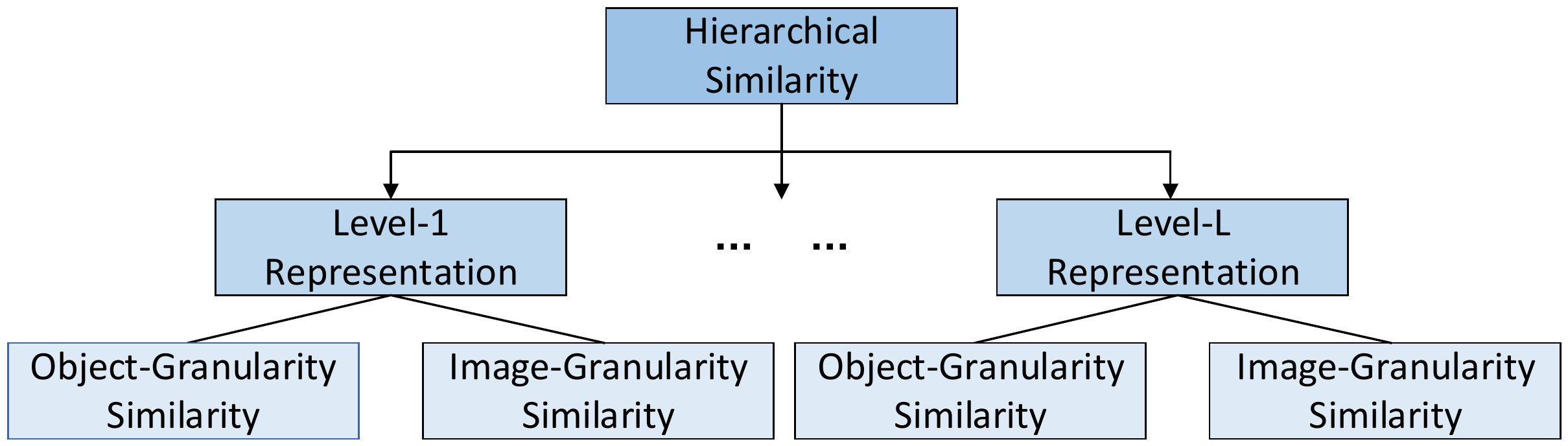}}\label{fig:concept_b} 
  \centerline{(b)}
\end{minipage}

\caption{(a) An example of language-based product image retrieval.
(b) Our proposed HSL learns cross-modal similarities between images and textual queries in a hierarchical manner.}
\vspace{-0.2in}
\label{fig:concept}
\end{figure}

Cross-modal retrieval is a classical task at the intersection between computer vision and natural language processing, and has been widely explored \cite{peng2017overview,dong2017cross,rasiwasia2010new,dong2019dual}. 
Recently, with the increasing popularity of e-commerce platforms~\cite{ma2020fine,vasileva2018learning,yang2019interpretable,chen2020image}, language-based product image retrieval attracts increasing attention~\cite{kddcup20201st,kddcup20208th,kddcup202010th}.
As exemplified in Fig. \ref{fig:concept}(a), given a textual query, the task is asked to retrieve images containing products that are specified by the given query.
In contrast to general cross-modal retrieval, product images seem to be more diverse. As shown in Fig. \ref{fig:concept}(a), a clothing product can be shown in isolation or on the mannequin; can be folded or not; can be displayed with a clear or complex background, showing the challenging characteristics of the language-based product image retrieval task.

Recent works for language-based product image retrieval tend to exploit strong image and textual query representations to tackle the problem~\cite{kddcup20201st,kddcup20208th,kddcup202010th}.
For instance, Huang \etal \cite{kddcup20201st} borrow the ideas of Modular Co-Attention Networks (MCAN) \cite{yu2019deep} and VisualBERT \cite{li2019visualbert}, taking advantage of the power of multi-head self-attention/transformer \cite{vaswani2017attention} to encode images and textual queries. 
Deriving from LXMERT \cite{tan2019lxmert}, Zhang \etal\cite{kddcup20208th} first encode images and textual queries by stacked self-attention modules and subsequently employ cross-modal guided attention to obtain robust image and query representations. In \cite{kddcup202010th}, Ding \etal also leverage transformer-based encoder to represent images and textual queries, but they formulate the retrieval as a multi-task problem where image captioning \cite{vinyals2015show} is integrated as an extra task to constrain the learned image representation.
Although the above methods utilize stacked encoders, only the high-level visual and textual representations are used for similarity measurement. Besides, they only perform cross-modal matching on single granularity, \eg image-query matching \cite{kddcup20201st,kddcup20208th}, image-word matching \cite{kddcup202010th}.
Considering product images are of diverse characteristics, we argue that such single-granularity similarity based on the representation of a specific level is sub-optimal. 

\begin{figure*}[!ht]
  \centering
  \includegraphics[width=1.85\columnwidth]{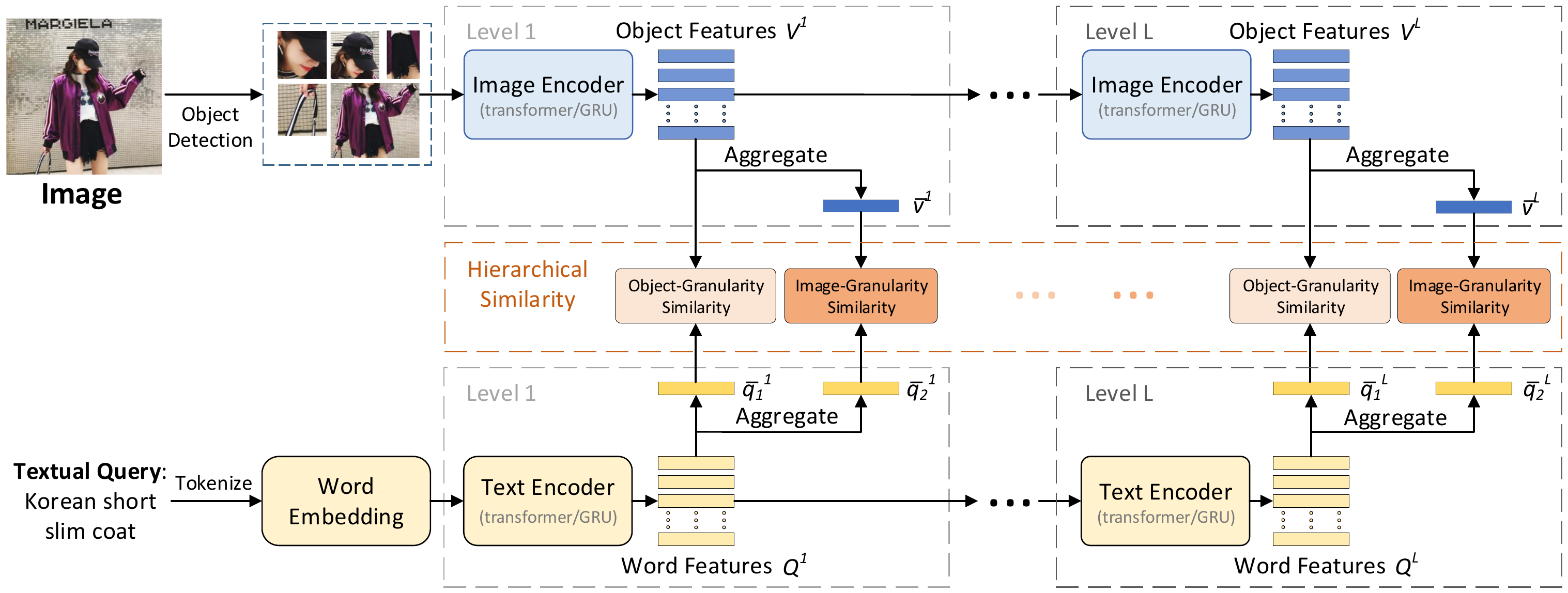}
  \vspace{-0.1in}
  \caption{Framework of our proposed Hierarchical Similarity Learning~(HSL) network.
  }
  \label{fig:framework}
  \vspace{-0.1in}
\end{figure*}

In this paper, we propose a Hierarchical Similarity Learning (HSL) network which simultaneously exploits multi-level representations of images and textual queries, and multi-granularity similarities at each level. As shown in Fig. \ref{fig:concept}(b), we exploit the multiple similarities in a hierarchical manner.
The framework of our proposed HSL is illustrated in Fig. \ref{fig:framework}. For both images and queries, multiple encoders are stacked to progressively learn the multi-level representations. These representations, generated by distinct encoders, are complementary to each other, which allows us to obtain an effective cross-modal similarity measurement. 
Moreover, considering product images usually include a number of objects, we propose to use both object-granularity similarity and image-granularity similarity to measure the relevance between images and queries.
In summary, the main contributions of this paper are:
\begin{itemize}
    \item We propose a Hierarchical Similarity Learning~(HSL) network, which jointly exploits multi-level representations and multi-granularity similarities. Based on the  multi-level representations and multi-granularity similarities, the final cross-modal similarity between images and queries is computed in a hierarchical manner.
    \item Our proposed HSL consistently performs better than general cross-modal retrieval models and models particularly designed for product image retrieval. Experiments on a large-scale language-based product image retrieval dataset demonstrate the effectiveness of our proposed method.
\end{itemize}

%% file: tex/method.tex
\section{Hierarchical Similarity Learning}

\subsection{Multi-level Representation}\label{sec:enc}
Given a product image $I$ and a textual query $T$, we first learn multi-level representations of image and text such that cross-modal similarities of different levels can be simultaneously learned. Through the multi-level encoding network, the image $I$ and the sentence $T$ are represented by a sequence of multi-level image features $\{ V^1, V^2, ..., V^L \}$ and a sequence of multi-level text features $\{ Q^1, Q^2, ..., Q^L \}$, $L$ is the number of total levels.

\subsubsection{Image Representation}
For the given product image $I$, a pre-trained object detection model is employed to detect main objects in the image, producing $n$ detected objects $V^0:\{v_k^0 \}_{k=1}^n$, where $v_k^0 \in \mathbb{R}^{d_0}$ is the extracted feature vector of $k$-th object. 
Before feeding them into the first encoder, we use a linear layer to project object features to $d_c$-dimensional feature vectors which fit the input size of the following encoders.
In order to obtain sufficient representation, we stack multiple encoders and utilize all their outputs as the multi-level representations.
More concretely, by employing an encoder on the initial object features $V^0:\{v_k^0 \}_{k=1}^n$, we obtain the level-1 representation:
\begin{equation}
    V^1 = \phi_1(V^{0}),
\end{equation}
where $\phi_1$ denotes the first image encoder. To capture the dependencies between objects, we utilize transformer~\cite{vaswani2017attention} as our fundamental encoder which has been found effective in various tasks~\cite{lu2019vilbert,devlin2018bert} due to its superior ability to model sequential relation. Note that other sequential models such as GRU~\cite{cho2014learning} or LSTM~\cite{hochreiter1997long} can also be used as the encoder.
Similarly, the outputs of the following encoders are represented by:
\begin{equation}
    V^l = \phi_l(V^{l-1}), l=2, ..., L,
\end{equation}
where $\phi_l$ indicates $l$-th image encoder, $L$ is the number of stacked encoders.
Finally, through $L$ stacked encoders, the multi-level representations of image $I$ are obtained as a sequence of feature groups $\{ V^l:\{v_k^l \}_{k=1}^n\}_{l=1}^L$.

\subsubsection{Text Representation}
Given a textual query $T$ of $m$ words, we first embed each word into a word vector space by GloVe word2vec~\cite{pennington2014glove}, resulting in a sequence of word features $Q^0:\{q_k^0\}_{k=1}^m$. Similar to the image encoder, a sequence of word features are then fed into a linear layer to change its dimension, followed by $L$ stacked encoders:
\begin{equation}
    Q^l = \psi_l(Q^{l-1}), l=1,2, ..., L,
\end{equation}
where $\psi_l$ denotes $l$-th text encoder.
Finally, through $L$ stacked encoders, the query $T$ can be represented as $\{ Q^l:\{q_k^l\}_{k=1}^m\}_{l=1}^L$.

\subsection{Multi-granularity Similarity}
Given both multi-level representations of images and textual queries, we propose to use multi-granularity similarity to measure their cross-modal similarity.
For image and query representations at each level, we compute their object-granularity similarity and image-granularity similarity respectively.
In what follows, we describe how to compute these two similarities based on the level-$l$ image representation $V^l:\{ v_k^l \}_{k=1}^{n}$ and query representation $Q^l:\{ q_k^l \}_{k=1}^{m}$.

\subsubsection{Object-granularity Similarity}
To compute the object-granularity similarity, we learn to map object and query features into a common object-query embedding space, where the object-query similarity can be directly measured. The final object-granularity similarity between an image and a query is obtained by aggregating the object-query similarities of all objects in the image.
Specifically, we first aggregate the word representations $\{ q_k^l \}_{k=1}^{m}$ of level $l$ into a query-level feature vector by mean pooling. A linear layer is further employed to project it into the object-query common embedding space, denoted as $\Bar{q}_1^l$.
For $n$ object features  $\{ v_k^l \}_{k=1}^{n}$ of level $l$ in the given image, we also employ a linear layer to transform them into the object-query embedding space, denoted by $\{\Bar{v}_k^l\}_{k=1}^n$.
Finally, the object-granularity cross-modal similarity between the image $I$ and query $T$ is computed as the average of similarities of all object-query pairs:
\begin{equation}\label{eq:object-level}
    \sigma_{obj}^l(I,T) = \frac{1}{n} \sum_{k=1}^{n} r(\Bar{v}_k^l, \Bar{q}_1^l),
\end{equation}
where $r(,)$ denotes the similarity function.
In our implementation, as we perform text-to-image retrieval, we utilize projection length of the textual query feature onto the visual feature~\cite{ying2018CMPM}:
\begin{equation}
     r(\Bar{v}_k^l, \Bar{q}_1^l) = \Bar{v}_k^l \cdot \Bar{q}_1^l / \Vert \Bar{v}_k^l \Vert.
\end{equation}

\subsubsection{Image-granularity Similarity}
As for image-granularity similarity, we focus on global similarity between images and queries.
Similarly, we map image and query representations to a common image-query embedding space.
For textual queries, we aggregate the word-level feature vectors $\{ q_k^l \}_{k=1}^{m}$ by mean pooling, followed by another linear layer to obtain the sentence-level feature $\Bar{q}_2^{l}$. Note that the linear layer here does not share weights with that in the object-granularity similarity computation, as we are actually learning two distinct embedding spaces for two similarities.
As product images generally contain abundant objects, it is necessary to capture salient features.
In such consideration, we employ an attention module to aggregate object features into an image-level feature vector.
Specifically, the attention weight for each object is computed by a multi-layer perceptron with one hidden layer. Formally, the attention weight for $k$-th object feature $v_k^l$ at level $l$ is computed as:
\begin{equation}
    w_k^l = W_2 \delta (W_1 v_k^l  + b_1) + b_2,
\end{equation}
where $\delta(\cdot)$ is the ReLU nonlinear activation, $W_1$, $W_2$ and $b_1$, $b_2$ denote the transformation matrices and biases respectively. Besides, we use a softmax layer to normalize the attention weights. 
With the learned attention weights $\{ w_k^l \}_{k= 1}^n$, the image-level feature is obtained as a weighted sum of all object features:
\begin{equation}
    \Bar{v}^l = \sum_{k=1}^{n} w_k^l v_k^l.
\end{equation}
Finally, we define the image-granularity similarity between the query $T$ and the image $I$ at level $l$ as:
\begin{equation}\label{eq:image-level}
    \sigma_{img}^l(I,T) = r(\Bar{v}^l,\Bar{q}_2^l).
\end{equation}

\subsection{Training and Evaluation}
\subsubsection{Model Training}
To train the model, we use the cross-modal projection matching loss~\cite{ying2018CMPM}, which aims to learn a common space where the similarity between relevant pairs are forced to be greater than irrelevant pairs in a contrastive manner. Different from \cite{ying2018CMPM} which only utilizes the loss over the final output of models, we employ the loss not only over the multi-level representations but also over the multiple granularities.
Specifically, at each level, we employ the loss on both object-granularity similarity $\sigma_{obj}^l$ and image-granularity similarity $\sigma_{img}^l$.
Formally, given a mini-batch with $N$ relevant image-query pairs, the loss on the object-granularity similarity at the level $l$ is:
\begin{equation}
    \mathcal{L}_{obj}^l = \frac{1}{N} \sum_{i=1}^{N} -log\frac{exp(\sigma_{obj}^l(I_i,T_i)}{\sum_{j=1}^{N} exp(\sigma_{obj}^l(I_i,T_j))},
\end{equation}
where $(I_i, T_i)$ indicates relevant image-query pair, $(I_i, T_j)$ denotes irrelevant image-query pair if $i \neq j$. Similarly, the loss on the image-granularity similarity is defined as:
\begin{equation}
    \mathcal{L}_{img}^l = \frac{1}{N} \sum_{i=1}^{N} -log\frac{exp(\sigma_{img}^l(I_i,T_i)}{\sum_{j=1}^{N} exp(\sigma_{img}^l(I_i,T_j))}.
\end{equation}
As we employ the loss over all the representations of distinct levels, the final overall loss of a model with $L$-level representations is:
\begin{equation}
    \mathcal{L} = \sum_{l=1}^L \lambda_l (\mathcal{L}_{obj}^l + \mathcal{L}_{img}^l).
\end{equation}
where $\lambda_l$ is a hyper-parameter of level $l$ which control the balance between different levels.

\subsubsection{Evaluation}
After the model being trained, we measure the similarity between product images and textual queries in terms of multi-level representations and multiple granularities.
Concretely, given a product image $I$ and a textual query $T$, their hierarchical similarity $s(I,T)$ is obtained by:
\begin{equation}
    s(I, T) = \sum_{l=1}^L \lambda_l (\sigma_{obj}^l(I, T)+\sigma_{img}^l(I, T)).
\end{equation}
With the hierarchical similarity, given a textual query, all candidate product images are ranked according to their similarities with the given query in descending order.

%% file: tex/exp.tex
\section{Experiments}
\subsection{Setup}
\subsubsection{Dataset and Metric}
We conduct experiments on the dataset of KDD Cup 2020 Challenges for Modern E-Commerce Platform: Multimodalities Recall~\cite{kddcup2020data}, a large-scale dataset for language-based product image retrieval.
The dataset is comprised of a training set of 3 million product images, a validation set of 9177 product images, and two test sets. As the ground-truth of the two test sets are not open-released, we report performance on the validation set.
Each training image is annotated with a textual phrase or a sentence which describes the specific product in the image.
In the validation set, there are 496 textual queries, and each query is along with a candidate pool of around 30 product images.

Following the evaluation protocol of the dataset, we use the nDCG@5 as the performance metric. We also report the performance of nDCG@k (k=10,15,20,25,30) to obtain a more comprehensive evaluation.

\subsubsection{Implementation Details}
We use PyTorch as our deep learning environment.
For object detection, we directly use the detected objects and extracted features provided by the dataset.
For the image encoder, we utilize a 3-layer transformer with 4 heads and hidden dimension of 512.
For textual query pre-processing, we first convert all words to the lowercase and then replace words that occur less than 5 times in the training set with a special token $\langle unk\rangle$. For the text encoder, we use a 2-layer transformer with the same structure as the image encoder. We also use one-layer bi-GRU as per-level encoder with dimension of 512. Both images and queries are embedded into space of 1024 dimension.
For the model loss, we empirically set $\lambda_1$ to 0.5 and $\lambda_2$ to 1.
To train the model, we utilize Adam optimizer\cite{kingma2014adam}. The initial learning rate is  set to 2e-4, decayed by 0.1 every epoch. The network is totally trained for 30 epochs. At each iteration, 256 query-product pairs are randomly sampled.

\subsection{Comparison with the State-of-the-art}

\input{tex/tab1}

\begin{figure}[!tb]
    \centering
    \includegraphics[width=0.7\columnwidth]{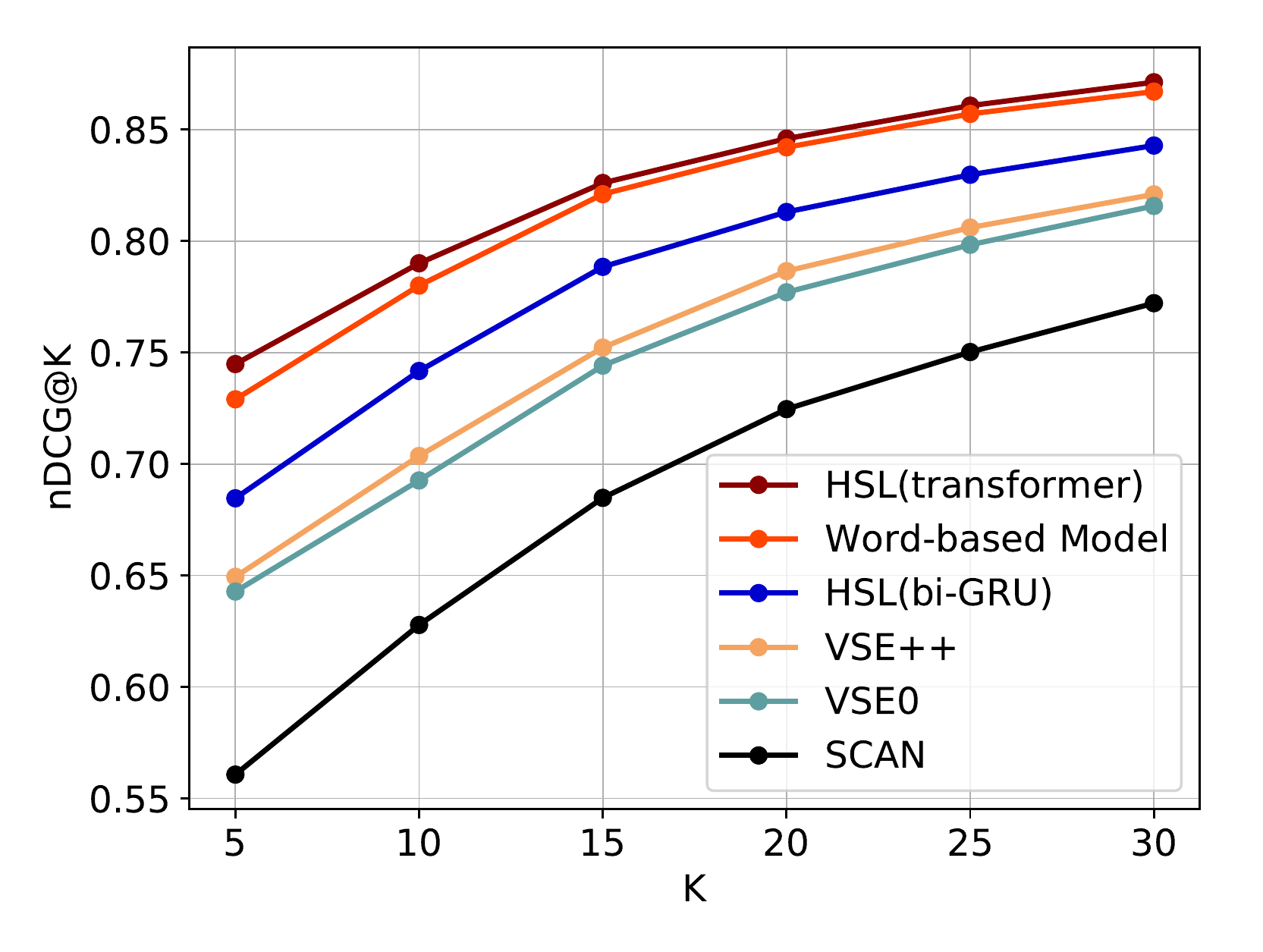}
    \vspace{-0.1in}
    \caption{Performances comparison in terms of varied $k$ of nDCG@k. Our HSL consistently performs the best.}
    \label{fig:ndcg-k}
    \vspace{-0.2in}
\end{figure}

To verify the viability of our proposed model, we compare it with two groups of works: one group consists of general image-text retrieval methods, the other are methods particularly designed for language-based product image retrieval. The results are summarized in Table \ref{tab:tab1}.
Our proposed model HSL with transformer encoder obtains nDCG@5 score of 0.7488, which outperforms the two groups of methods with a clear margin. The result shows the effectiveness of HSL for language-based product image retrieval. 

Among the first group, SCAN measures the image-text similarity in terms of the object granularity, while both VSE0 and VSE++ only consider the image-granularity similarity.
As these three models utilize GRU-based text encoding, we replace the transformer encoders in HSL with bidirectional GRU (bi-GRU) to make the comparison fairer. 
HSL equipped with bi-GRU outperforms these three methods with a clear margin. It shows the importance of multi-granularity similarity for product image retrieval. 
Among the second group, all models employ transformers to encode input data, but they only consider specific granularity similarity at high level. The better performance of HSL verifies the effectiveness of our hierarchical similarity learning framework.
Additionally, we also report $nDCG@k$ of varied $k$ in Fig. \ref{fig:ndcg-k}. Our HSL consistently outperforms the other counterparts.

\begin{figure}[!tb]
  \centering
  \includegraphics[width=0.75\columnwidth]{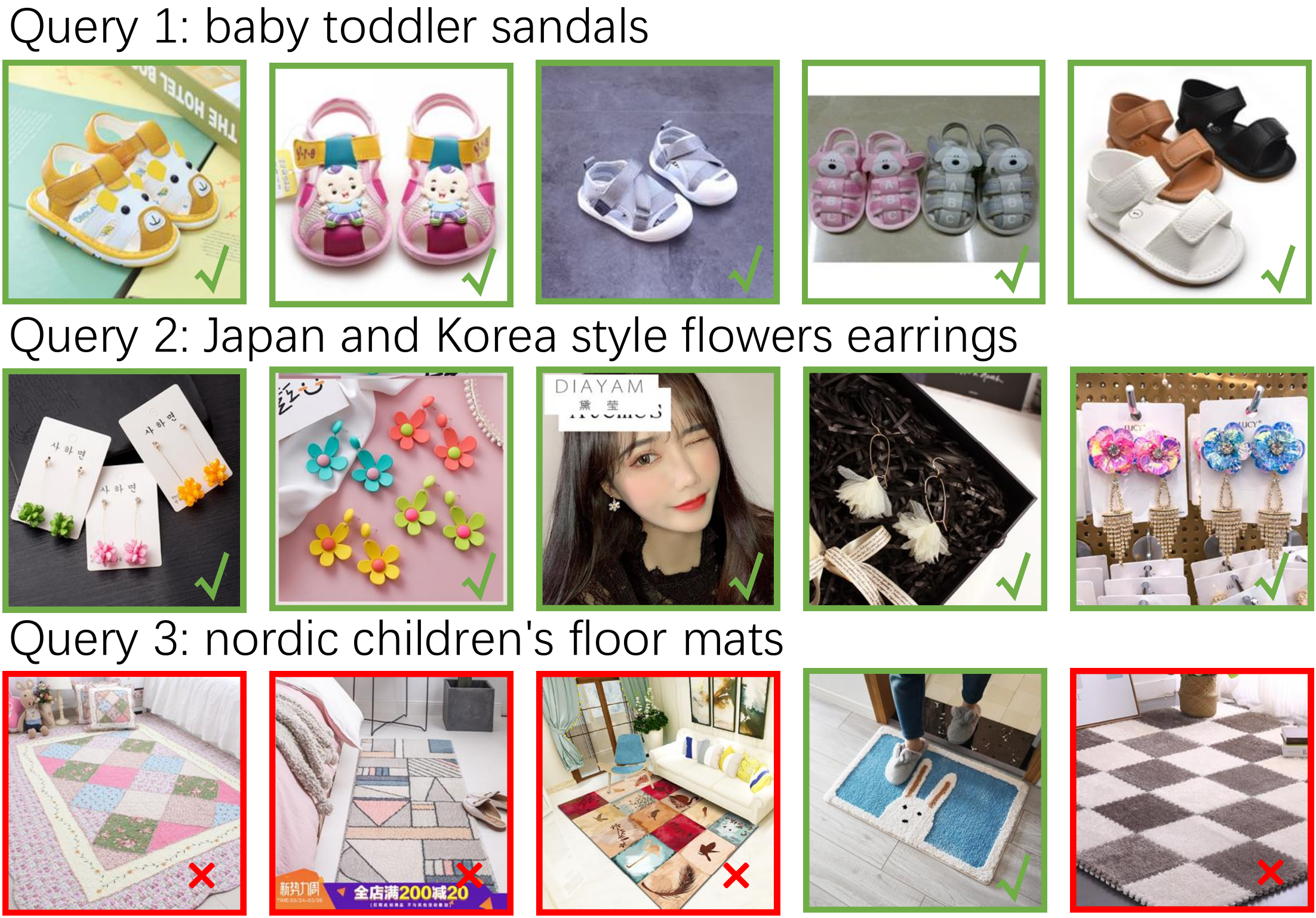}
  \vspace{-0.1in}
  \caption{For each textual query, top-5 product images sorted in descending order of similarity are presented. Green bounding box indicates correct one, while red ones are incorrect. }
  \label{fig:retrieval_examples}
  \vspace{-0.2in}
\end{figure}

Fig. \ref{fig:retrieval_examples} displays some qualitative results of our proposed HSL. 
HSL performs well for top 2 queries, while bad for Query 3 of \textit{nordic children's floor mats}.
Although top-5 retrieved images of Query 3 are all mat products, our model can not distinguish which mats are for children and which are not thus give the relatively bad result.

\subsection{Ablation Study}

\input{tex/tab2}

Table \ref{tab:tab2} summarizes the results of the ablation study. It can be seen that HSL suffers the performance degeneration when a single-level representation is used, which shows the effectiveness of our multi-level representations.
Among the similarity granularity, using the object-granularity similarity performs better than the image-granularity one, and multi-granularity similarities considering both granularities achieve the best result. The result shows the complementarity of the object-granularity similarity and the image-granularity similarity. 
Moreover, our full model with both multi-level representations and multi-granularity similarities gives the best nDCG@5 score of 0.7448, which further verifies the importance of our proposed hierarchical similarity learning for language-based product image retrieval.
We also try HSL with more than two levels, while find that using more levels does not lead to better performance.

%% file: tex/tab1.tex
 \begin{table}
 \centering
   \caption{Performance comparison with state-of-the-art models. Our proposed HSL with the transformer as encoders performs the best. For a fair comparison, all the scores are obtained by single model without model ensemble.}
   \label{tab:tab1}
   \scalebox{0.95}{
   \begin{tabular}{lr}
     \toprule
      Method & nDCG@5  \\
     \midrule
     SCAN\cite{lee2018stacked} & 0.5609  \\
     VSE0\cite{faghri2018vse++}& 0.6381  \\
     VSE++\cite{faghri2018vse++}& 0.6494  \\
     \midrule
     LXMERT+LightGBM \cite{kddcup20208th} &  0.6200  \\
     MCAN\cite{yu2019deep,kddcup20201st} &  0.6900  \\
     VisualBERT\cite{li2019visualbert,kddcup20201st} & 0.7100  \\
     Word-based Model\cite{kddcup202010th} & 0.7290 \\
     \midrule
     HSL(bi-GRU) & 0.6846 \\
     HSL(transformer) & \textbf{0.7448}\\
   \bottomrule
 \end{tabular}
 }
 \end{table}

%% file: tex/tab2.tex
\begin{table}
 \centering
   \caption{Ablation study of HSL.  The \XSolidBrush symbol indicates model without multiple-level representation or multi-granularity similarity, and    used single level or granularity are specified in parenthesis. Our full model performs the best.}
   \label{tab:tab2}
   \scalebox{0.85}{
   \begin{tabular}{ccr}
     \toprule
     Multi-level Representation? & Multi-granularity Similarity ? & nDCG@5  \\
    \midrule
    \multirow{3}{*}{ \XSolidBrush (Level 1)} & \XSolidBrush (Object) & 0.7275\\
    & \XSolidBrush (Image) & 0.7240\\
    & \Checkmark & 0.7339\\
    \midrule
    \multirow{2}{*}{ \XSolidBrush (Level 2)} & \XSolidBrush (Object) & 0.7351\\
    & \XSolidBrush (Image) & 0.7273\\
    & \Checkmark & 0.7362\\
    \midrule
    \multirow{3}{*}{\Checkmark}& \XSolidBrush (Object) & 0.7412\\
    & \XSolidBrush (Image) & 0.7381\\
    & \Checkmark & \textbf{0.7448}\\
   \bottomrule
 \end{tabular}
 }
 \vspace{-0.2in}
 \end{table}

%% file: tex/conclusion.tex
\section{Conclusion}
This paper proposes a hierarchical similarity learning network for language-based product image retrieval task. Different from existing works that consider single-granularity similarity based on the representation of a specific level, we compute multi-granularity similarities based on multi-level representations in a hierarchical manner. Experiments on a large-scale product image retrieval dataset verify the viability of our model for language-based product image retrieval, and the ablation study shows the importance of both multi-level representations and multi-granularity similarities. 
In the future, we would like to explore our model for general cross-modal retrieval tasks, such as text-to-video retrieval~\cite{li2019w2vv++,yang2020tree}.